\documentclass[journal,twoside,web]{ieeecolor}
\providecommand{\refname}{REFERENCES}
\usepackage{generic}
\usepackage{cite}
\usepackage{amsmath,amssymb,amsfonts}
\usepackage{algorithmic}
\usepackage{graphicx}
\usepackage{algorithm,algorithmic}
\usepackage{hyperref}
\hypersetup{hidelinks}
\usepackage{textcomp}

\usepackage{booktabs}
\usepackage{array}
\usepackage{multirow}
\usepackage{multicol}
\usepackage{cleveref}
\crefname{figure}{Fig.}{Figs.}
\Crefname{figure}{Fig.}{Figs.}
\crefname{table}{Table}{Tables}
\Crefname{table}{Table}{Tables}
\crefname{section}{Section}{Sections}
\Crefname{section}{Section}{Sections}

\def\BibTeX{{\rm B\kern-.05em{\sc i\kern-.025em b}\kern-.08em
    T\kern-.1667em\lower.7ex\hbox{E}\kern-.125emX}}
\begin{document}
\title{CoMLP: Cooperatively-Gated MLPs for Fine-Grained Cross-Modal Information Fusion \\in Medical Image Segmentation}

\author{Mingyuan Meng, Shuchang Ye, Mingjian Li, Zhenyu Zhao, Jinman Kim, and Lei Bi
\thanks{This work was supported by the National Natural Science Foundation of China under Grant No. 82595964, the Fundamental Research Funds for the Central Universities under Grants No. YG2025ZD26/JC202622, and the Zhongguancun Academy under Grant XTS0071. (Corresponding authors: Jinman Kim and Lei Bi.)}
\thanks{Mingyuan Meng is with the Institute of Translational Medicine, Shanghai Jiao Tong University, Shanghai, China, and also with the Zhongguancun Academy \& Zhongguancun Institute of Artificial Intelligence, Beijing, China (mengmingyuan@bza.edu.cn).}
\thanks{Shuchang Ye, Mingjian Li, and Jinman Kim are with the School of Computer Science, the University of Sydney, Sydney, NSW, Australia (shuchang.ye@sydney.edu.au; limingjianbd@gmail.com; jinman.kim@\\sydney.edu.au).}
\thanks{Zhenyu Zhao is with China United Network Communications Group Co., Ltd., Beijing, China (zhaozy185@chinaunicom.cn).}
\thanks{Lei Bi is with the Institute of Translational Medicine, Shanghai Jiao Tong University, Shanghai, China, and also with the School of Computer Science, the University of Sydney, Australia (lei.bi@sjtu.edu.cn).}
}

\IEEEaftertitletext{\vspace{-1.5\baselineskip}}
\maketitle

\begin{abstract}
Multi-modal medical images and clinical reports provide complementary anatomical, functional, and semantic information for medical image segmentation. Effectively exploiting these heterogeneous sources requires fine-grained cross-modal information fusion that preserves subtle spatial details while capturing semantic dependencies across modalities. Existing fusion approaches frequently rely on cross-attention, whose computational burden increases rapidly with spatial resolution, making dense cross-modal interaction difficult on high-resolution feature maps, particularly for volumetric medical images. In this work, we propose CoMLP, a cooperatively-gated MLP module for fine-grained cross-modal information fusion in medical image segmentation. CoMLP models cross-modal dependencies through cooperative cross-gating, built upon complementary regional and dilated MLP interactions, to capture local and global cross-modal dependencies. We further develop a multi-source fusion architecture in which CoMLP performs both inter-image fusion across imaging modalities and vision-language fusion between visual features and textual reports, enabling heterogeneous information to be integrated without relying on dense cross-attention. Extensive experiments on five medical segmentation benchmarks, covering 2D/3D images, clinical reports, multiple imaging modalities, and diverse anatomical regions, demonstrate consistent improvements over state-of-the-art multi-modal and language-guided segmentation methods. Ablation studies further show that fine-grained interaction at high spatial resolutions and complementary local-global fusion are critical to the performance gains. These results demonstrate the potential of MLP-based interaction as an effective alternative for fine-grained cross-modal information fusion in medical image segmentation.
\end{abstract}


\section{Introduction}
\label{sec:intro}
\IEEEPARstart{M}{edical} image segmentation is a fundamental component of medical image analysis, supporting the delineation of anatomical structures and pathological lesions for diagnosis, treatment planning, and quantitative assessment \cite{ma2024segment}. Increasingly, accurate segmentation relies on heterogeneous information sources that characterize the patient from different perspectives. Multi-modal medical images provide complementary visual information; for example, CT depicts anatomical structures and tissue morphology, whereas PET reveals functional and metabolic activity \cite{xue2024multi}. Clinical reports provide another complementary source by describing high-level semantic information such as lesion location, morphology, extent, and imaging characteristics \cite{li2023lvit,huemann2026contextual}. Effectively exploiting these heterogeneous sources therefore requires cross-modal information fusion across substantially different representation spaces. In particular, medical image segmentation demands not only the integration of complementary information, but also the preservation of fine-grained spatial details that determine subtle lesion boundaries and anatomical structures.

Existing studies on such cross-modal information fusion have mainly developed along two related directions. The first focuses on inter-image fusion, where information from different medical imaging modalities is integrated to improve segmentation \cite{xue2021multi,podobnik2023multimodal,zheng2025asymmetric}. Recent PET/CT segmentation methods, for instance, explicitly model interactions between anatomical and metabolic imaging features through cross-modal transformers or other interaction mechanisms \cite{li2024swincross,huang2025c2maot,mei2025cross}. The second direction introduces clinical language as semantic guidance for segmentation. Beginning with language-guided medical image segmentation \cite{li2023lvit}, subsequent studies have developed increasingly sophisticated mechanisms to fuse visual features with textual descriptions \cite{zhong2023ariadne,lee2023text,guo2024common,zhang2024madapter,yu2025frequency,zeng2025harnessing,ye2024enabling,ye2025alleviating,han2026localization,li2026language}. More recently, vision-language modeling has also been extended to volumetric PET/CT, where textual descriptions are used to ground reported findings to corresponding lesions \cite{huemann2026contextual}. These advances demonstrate the value of both inter-image and vision-language information fusion. Nevertheless, the two types of fusion are commonly addressed using different mechanisms, while a general approach for fine-grained interaction across heterogeneous visual and textual information remains insufficiently explored.

This issue is particularly important for segmentation because the required fusion granularity is fundamentally different from that of image-level multi-modal prediction. Accurate delineation depends on local tissue appearance and boundary details, while also requiring broader contextual information to distinguish target structures from visually similar regions. For inter-image fusion, complementary information from different imaging modalities should therefore interact at sufficiently fine spatial scales so that subtle modality-specific characteristics are retained. For vision-language fusion, high-level textual semantics should likewise be selectively associated with spatial visual representations rather than simply introduced as global context. Consequently, an effective fusion mechanism should capture both local and long-range cross-modal dependencies while retaining high-resolution spatial information.

Cross-attention has been a predominant solution for cross-modal dependency modeling as it explicitly establishes interactions between features from different information sources \cite{petit2021u}. It has consequently been widely adopted for both multi-modal medical imaging and vision-language fusion \cite{meng2023merging,zhang2024madapter,guo2024common,zeng2025harnessing}. However, the computational and memory demands of cross-attention increase substantially as the number of interacting tokens grows. This issue is particularly pronounced for inter-image fusion, where two high-resolution spatial feature maps introduce large sets of visual tokens and dense pairwise interactions between them. In practice, cross-modal attention is therefore often performed after patch embedding, pooling, or other spatial reduction operations, particularly for 3D medical images \cite{meng2023merging,li2024swincross}. Hence, the challenge is not whether cross-attention can model cross-modal dependencies, but whether dense attention permits these dependencies to be modeled at the spatial granularity required by high-resolution medical image segmentation. This motivates an alternative fusion paradigm that avoids constructing dense pairwise attention while retaining explicit cross-modal interaction.

Multi-layer Perceptrons (MLPs) provide a promising direction toward this goal. MLP-based vision models, such as MLP-Mixer\cite{tolstikhin2021mlp} and gMLP\cite{liu2021pay}, have demonstrated that long-range visual dependencies can be captured without self-attention, and recent medical vision studies further show their effectiveness for fine-grained modeling of high-resolution image features \cite{meng2023full,meng2024correlation,li20243dpx,xue2026hybrid}. The potential of MLPs has also been explored in vision-language modeling. Recent vision-language frameworks employ MLP-Mixer modules primarily for feature projection, compression, or alignment \cite{xin2026med3dvlm,xing2026medvlsam2}, establishing the feasibility of MLPs in multi-modal modeling while leaving a question largely open: \textit{whether MLPs can serve as dense cross-modal interaction operators that explicitly capture fine-grained dependencies between heterogeneous information sources.} Such an operator would be particularly attractive if it could support both inter-image and vision-language fusion using a common formulation.

To this end, we propose CoMLP, a cooperatively-gated MLP module for fine-grained cross-modal information fusion in medical image segmentation. Rather than employing MLPs merely for feature projection or global conditioning, CoMLP introduces information from one modality into the dense interaction of another through cooperative cross-gating. It combines regional and dilated MLP interactions to capture complementary local and global cross-modal dependencies without relying on dense cross-attention. Based on CoMLP, we construct a multi-source medical information fusion architecture in which the same interaction primitive is used for two heterogeneous fusion processes: inter-image fusion between medical imaging modalities and vision-language fusion between visual features and clinical reports. In the former, CoMLP enables cross-modal interaction beginning from high-resolution visual feature maps; in the latter, it propagates report-derived semantics into multi-scale visual representations for segmentation. Our main contributions are summarized as follows:
\begin{itemize}
    \item We investigate MLP-based dense cross-modal interaction operators for fine-grained medical information fusion, providing an alternative to the prevailing attention-dominated paradigm for integrating heterogeneous visual and textual information.
    
    \item We propose CoMLP, which performs cooperative cross-gating with complementary regional and dilated MLP-based interactions to capture both local and global cross-modal dependencies at fine spatial granularity.
    
    \item We develop a CoMLP-based multi-source fusion architecture in which the same interaction primitive performs both inter-image and vision-language fusion.
\end{itemize}

Extensive experiments on five medical image segmentation benchmarks involving 2D/3D medical images, clinical reports, multiple imaging modalities, and diverse anatomical regions demonstrate its consistent effectiveness over state-of-the-art multi-modal and language-guided segmentation methods.

\section{Related Work}
\subsection{Cross-Modal Fusion of Multi-Modal Medical Images}
Multi-modal medical images provide complementary information for characterizing anatomical structures and pathological abnormalities, motivating extensive research on their fusion for medical image segmentation \cite{xue2024multi,wu2024review}. Early studies commonly integrated different imaging modalities by input concatenation or feature aggregation, while subsequent studies increasingly focused on explicitly modeling inter-modal interactions. For example, Xue et al. \cite{xue2021multi} proposed a co-learning framework to exploit complementary information between PET and CT for liver lesion segmentation, while Podobnik et al. \cite{podobnik2023multimodal} investigated multi-modal CT and MR fusion for head-and-neck organ segmentation. More recently, Zheng et al. \cite{zheng2025asymmetric} developed an asymmetric heterogeneous network to adaptively exchange information across imaging modalities.

Recent studies have further explored explicit cross-modal dependency modeling for multi-modal medical image segmentation. SwinCross \cite{li2024swincross} adopted cross-modal Swin Transformers to exchange information between PET and CT representations. C\textsuperscript{2}MAOT \cite{huang2025c2maot} employed cross-modal masked modeling and optimal transport to learn complementary modality information, while CIPA \cite{mei2025cross} introduced interactive perception with Mamba for tumor segmentation. These methods demonstrate the importance of inter-image information interaction beyond simple feature aggregation. Nevertheless, their fusion mechanisms are designed for interactions between images and are generally distinct from those developed for integrating vision-language information. This motivates a more general cross-modal mechanism that can accommodate heterogeneous information sources within a common formulation.

\subsection{Vision-Language Fusion for Medical Segmentation}
Clinical language provides high-level semantic information that is complementary to visual context in medical images. LViT \cite{li2023lvit} pioneered language-guided medical image segmentation by explicitly incorporating textual descriptions into segmentation, establishing QaTa-COV19 and MosMedData+ as commonly used vision-language segmentation benchmarks. Subsequent studies have developed increasingly sophisticated mechanisms for visual-textual information fusion. Ariadne's Thread \cite{zhong2023ariadne} integrated textual representations with multi-scale visual features via cross-attention, while cross-position attention \cite{lee2023text} and common vision-language attention \cite{guo2024common} are also leveraged to explicitly model semantic correspondences between visual and textual features. More recently, MAdapter \cite{zhang2024madapter} introduced bidirectional vision-language interaction, FMISeg \cite{yu2025frequency} performed textual interaction with visual features in the frequency domain, and TeViA \cite{zeng2025harnessing} enhanced text-to-vision alignment through segmentation-oriented semantic constraints.

Cross-attention remains a prevalent mechanism due to its ability to explicitly associate vision-language representations, although alternative mechanisms have also emerged. For example, ViTexNet \cite{bhardwaj2025vitexnet} converts textual information into dynamic convolutional guidance rather than performing conventional cross-attention. Language-guided segmentation has also begun to extend from uni-modal settings toward multi-modal imaging. ConTEXTual Net 3D \cite{huemann2026contextual}, for instance, concatenates PET/CT as the visual input, without explicit inter-image interaction, and subsequently employs cross-attention to associate lesion descriptions with visual representation for volumetric grounding. These advances demonstrate the growing importance of vision-language fusion in medical image segmentation. However, image-text interaction is still generally separate from inter-image fusion, and a common mechanism capable of fine-grained interaction across both types of heterogeneous information remains comparatively underexplored.

\subsection{MLP-based Cross-Modal Modeling}
MLP-based models provide an alternative to self-attention for long-range visual dependency modeling. MLP-Mixer \cite{tolstikhin2021mlp} and gMLP \cite{liu2021pay} demonstrated that token interactions can be modeled using MLP operations without relying on self-attention. This property has subsequently been exploited in medical image analysis, where fine-grained medical detail information is particularly important. Recent studies have demonstrated the effectiveness of MLPs for long-range dependency modeling on high-resolution medical image features, widely benefiting diverse dense prediction tasks in medical imaging including segmentation \cite{meng2023full}, registration \cite{meng2024correlation}, reconstruction \cite{li20243dpx}, and synthesis \cite{xue2026hybrid}. MLPs have also been explored for feature interaction. MAXIM \cite{tu2022maxim} introduces multi-axis MLPs together with cross-gating for cross-feature conditioning. Similar MLP-based cross-gating has subsequently been explored in MCNMF-Unet \cite{yuan2024mcnmf} and MACG-Net \cite{yuan2024macg} for feature interaction in medical image segmentation and registration. These studies differ from ours in three aspects. First, conventional cross-gating generates the modulation of one feature stream from the complementary stream, whereas CoMLP cooperatively constructs the gate using both the mainstream and cross-modal information. Second, existing cross-gating designs focus on interactions between visual features and have not been generalized to heterogeneous vision-language fusion. More importantly, they primarily focus on the interaction operator itself rather than explicitly investigating the spatial granularity of feature interaction, which is particularly important for medical image segmentation where fine-grained cross-modal dependencies need to be modeled with subtle details.

MLPs have also been introduced into vision-language modeling. MLP-ViL \cite{nie2021mlp} empirically investigated MLP-based token mixing for general vision-language fusion, demonstrating the potential of MLPs for multi-modal learning. In medical vision-language modeling, Med3DVLM \cite{xin2026med3dvlm} employs a dual-stream MLP-Mixer projector to integrate multi-level visual representations with language embeddings, while MedVL-SAM2 \cite{xing2026medvlsam2} uses MLP-based projection to connect visual representations with a language model for multi-modal reasoning. These studies establish the feasibility of MLPs for multi-modal modeling; however, MLPs are predominantly used for representation projection, token mixing, or global conditioning, rather than as dense cross-modal interaction operators for explicitly modeling fine-grained dependencies on high-resolution feature maps. Our CoMLP investigates this role and further employs the same MLP-based interaction primitive for both inter-image and vision-language information fusion.

\section{Method}
\label{sec:method}
\label{sec:overview}
We formulate medical image segmentation with heterogeneous information sources as a fine-grained cross-modal information fusion problem. Given medical images \(I_{M_1}\) and \(I_{M_2}\) acquired from two complementary imaging modalities and a corresponding clinical report \(T\), our objective is to predict a segmentation mask \(Y\) by jointly exploiting modality-specific visual information and report-derived semantic information. This involves two distinct forms of cross-modal interaction: inter-image fusion between \(I_{M_1}\) and \(I_{M_2}\), and vision-language fusion between visual representations and textual semantics.

To accommodate these heterogeneous interactions within a common formulation, we introduce CoMLP as a general cross-modal interaction operator:
\begin{equation}
    \hat{F}_{\mathrm{main}}
    =
    \Phi_{\mathrm{CoMLP}}
    \left(
    F_{\mathrm{main}},
    F_{\mathrm{cross}}
    \right),
    \label{eq:comlp_general}
\end{equation}
where \(F_{\mathrm{main}}\) denotes the representation to be enhanced and \(F_{\mathrm{cross}}\) provides complementary information from another modality. CoMLP is built upon three key designs. First, a cooperative cross-gating unit explicitly introduces \(F_{\mathrm{cross}}\) into the spatial gating of \(F_{\mathrm{main}}\). Second, regional and dilated MLP interactions capture both local and global cross-modal dependencies. Third, lightweight modality adaptation allows the same interaction primitive to operate on both spatially aligned image features and heterogeneous image-text representations.

In the following sections, we first detail the core CoMLP mechanism and its instantiation across heterogeneous modalities, followed by a CoMLP-based multi-source fusion architecture for multi-modal segmentation.

\begin{figure*}[htb]
  \centering
  \includegraphics[height=7.5cm]{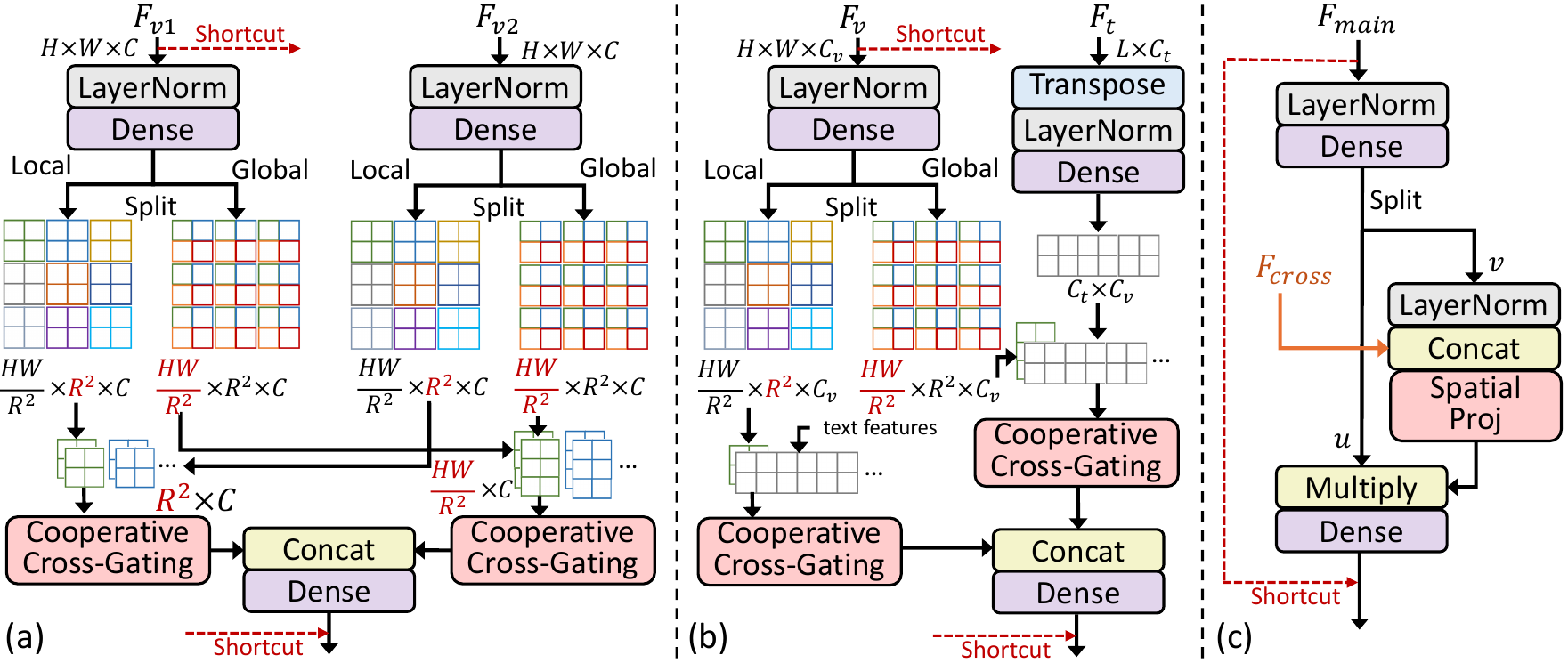}
  \caption{Illustration of CoMLP for fine-grained cross-modal information fusion. (a) CoMLP instantiated for inter-image fusion between two imaging modalities. (b) CoMLP instantiated for vision-language fusion between visual and textual representations. (c) Cooperative cross-gating unit where cross-modal information participates in gate generation to modulate the mainstream representation.}
  \label{fig:comlp}
\end{figure*}

\subsection{CoMLP: Fine-Grained Cross-Modal Interaction}
\label{sec:CoMLP}
As illustrated in \cref{fig:comlp}, CoMLP embeds cooperative cross-gating units into a dual-path structure inspired by multi-axis feature processing \cite{zhao2021improved}. Instead of conducting pairwise attention between cross-modal features, CoMLP performs MLP-based interactions along complementary local and global spatial axes, enabling high-resolution representations to directly participate in fine-grained dense cross-modal interaction.

\subsubsection{Cooperative Cross-Gating}
\label{sec:gating}
The central operation of CoMLP is the cooperative cross-gating unit, which builds upon the gating principle of gMLP \cite{liu2021pay} and jointly incorporates cross-modal information into gate generation. Given a mainstream representation \(F_{\mathrm{main}}\), we first apply feature projection and split the projected representation into two feature components:
\begin{equation}
    [u,v]
    =
    \operatorname{Split}
    \left(
    \mathcal{P}
    \left(F_{\mathrm{main}}\right)
    \right),
    \label{eq:split}
\end{equation}
where \(\mathcal{P}(\cdot)\) denotes a learnable feature projection and \(u\) and \(v\) have identical dimensions. In conventional gMLP, the spatial gate applied to \(u\) is generated solely from \(v\). Here, we instead concatenate \(v\) with the cross-modal representation \(F_{\mathrm{cross}}\) and jointly use them to construct the gating response:
\begin{equation}
    g
    =
    \mathcal{S}
    \left(
    \mathcal{D}
    \left(
    [v;F_{\mathrm{cross}}]
    \right)
    \right),
    \label{eq:crossgate}
\end{equation}
where \([\cdot;\cdot]\) denotes feature concatenation, \(\mathcal{D}(\cdot)\) performs cross-modal feature integration, and \(\mathcal{S}(\cdot)\) denotes spatial projection along the interaction axis. The mainstream response is then modulated through:
\begin{equation}
    \hat{u}
    =
    u \odot g,
    \label{eq:gating}
\end{equation}
where \(\odot\) denotes element-wise multiplication.

The key distinction from conventional cross-gating \cite{tu2022maxim} is therefore that the spatial gate is cooperatively determined by both information sources:
\begin{equation}
    g =
    g(F_{\mathrm{main}},F_{\mathrm{cross}}).
\end{equation}

Consequently, cross-modal information is not merely appended to the output representation or used to generate a global conditioning vector; it directly participates in determining which spatial responses of the mainstream feature should be emphasized or suppressed.

\begin{figure*}[htb]
  \centering
  \includegraphics[height=7cm]{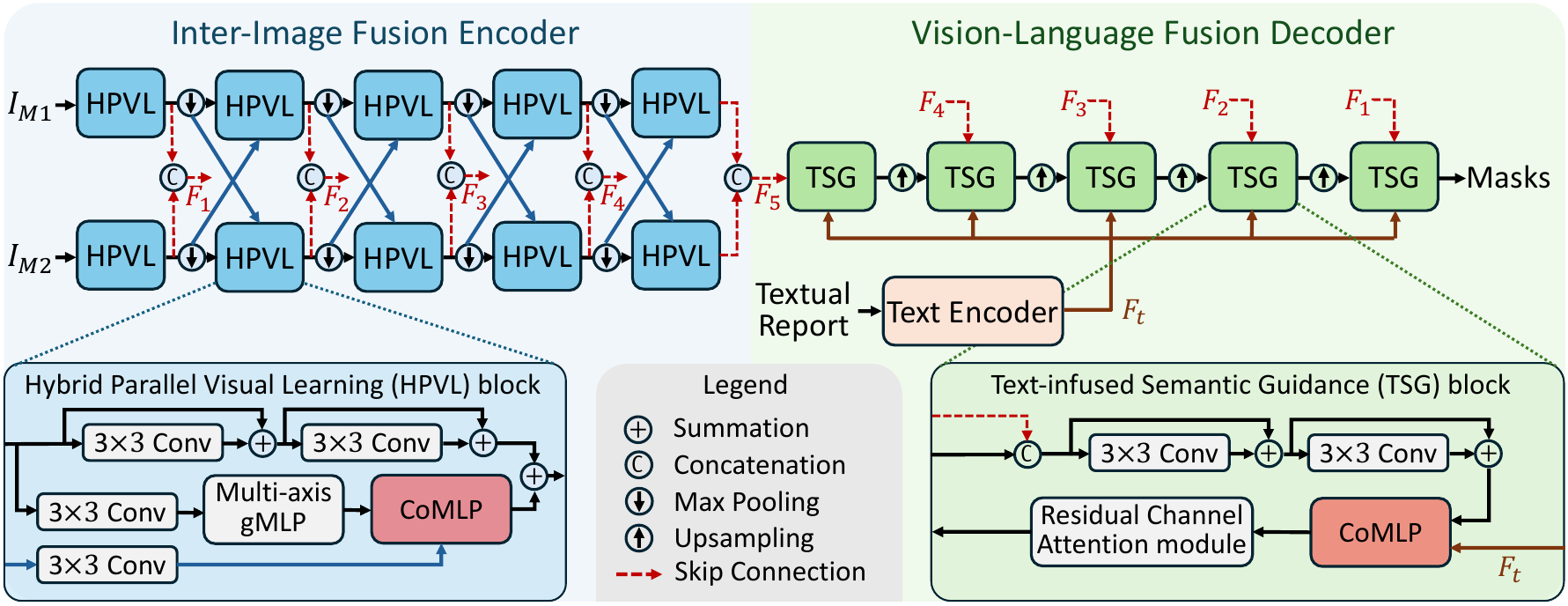}
  \caption{Illustration of CoMLP-based multi-source fusion architecture. CoMLP is applied throughout the encoder for inter-image fusion between medical imaging modalities and in the decoder for vision-language fusion between visual and textual representations.}
  \label{fig:architecture}
\end{figure*}

\subsubsection{Local-Global Multi-Axis Interaction}
\label{sec:localglobal}
Medical image segmentation requires both subtle local details and long-range global contexts. Directly performing unrestricted global interaction on high-resolution features, however, can result in substantial computational/memory demands and does not explicitly distinguish fine local dependencies from long-range contextual interactions. We hence factorize cross-modal interaction into complementary regional and dilated branches for both local and global feature interaction.

For a \(d\)-dimensional visual feature map, where \(d=2\) or \(3\), let the number of spatial elements be:
\begin{equation}
    N=\prod_{i=1}^{d}S_i,
\end{equation}
where \(S_i\) denotes the spatial size along the \(i\)-th dimension. Given a region size \(R\), each local region contains \(P=R^d\) spatial elements, resulting in \(G=\frac{N}{P}\) regions. The visual representation can therefore be reorganized into \(F^{r}\in\mathbb{R}^{G\times P\times C}\).

In the regional branch, a cooperative cross-gating unit operates along the \(P\)-dimensional intra-region axis. Spatially neighboring features within each region therefore participate in cross-modal gating, thus capturing subtle local cross-modal dependencies between modality-specific details. Such interactions are particularly useful for integrating modality-specific tissue appearance, lesion texture, and boundary information.

In the dilated branch, the representation is reorganized such that features at corresponding locations across local regions are interacted along the \(G\)-dimensional axis. A cooperative cross-gating unit is then applied across these locations, enabling long-range global cross-modal dependencies to be modeled across local regions. Such interactions are particularly useful for providing global contextual awareness to capture spatially distant yet semantically correlated cross-modal cues.

CoMLP comprises regional and dilated branches for complementary enhancement, where the former emphasizes subtle local dependencies and the latter captures long-range global dependencies. Their outputs are restored to the original spatial shape, and then concatenated and projected to form the cross-modally enhanced representation.

By avoiding the construction of dense pairwise attention matrices, MLP-based interaction itself provides a computationally favorable alternative for high-resolution dependency modeling, as demonstrated by previous MLP-based vision studies \cite{tolstikhin2021mlp,liu2021pay,meng2023full}. CoMLP further reduces the interaction burden by factorizing cross-modal dependency modeling along structured regional and dilated axes, rather than performing unrestricted spatial mixing over the entire feature map. Together, these designs make cross-modal fusion practically feasible at higher spatial resolutions under a limited computational and memory budget, enabling feature interaction at higher resolutions with less detail degradation and therefore facilitating finer-grained cross-modal information fusion.

\subsubsection{Unified Interaction across Heterogeneous Modalities}
\label{sec:heterogeneous}
Although inter-image and vision-language fusion involve substantially different information sources, CoMLP treats them under the same main-cross interaction formulation in \cref{eq:comlp_general}. Modality-specific adaptation is only used to prepare heterogeneous representations for cooperative gating.

\textbf{Inter-image interaction.}
For two medical image features \(F_{v1},F_{v2}\in\mathbb{R}^{H\times W\times C}\) (or \(\mathbb{R}^{H\times W\times D\times C}\) for 3D data), the two representations share a common spatial coordinate system but encode complementary imaging characteristics. For example, CT primarily characterizes anatomical morphology whereas PET represents metabolic activity. Within each modality-specific branch, the feature from the complementary modality serves as \(F_{\mathrm{cross}}\) and cooperatively modulates the mainstream feature \(F_{\mathrm{main}}\). Both representations are processed through the regional and dilated interaction branches described above, enabling local and global inter-image dependency modeling at the corresponding spatial locations.

\textbf{Vision-language interaction.}
Visual and textual representations have more heterogeneous structures. Given a visual feature \(F_v\in\mathbb{R}^{H\times W\times C_v}\) and a text feature \(F_t\in\mathbb{R}^{L\times C_t}\), where \(L\) is the textual sequence length and \(C_t\) is the language embedding dimension, we first transpose the textual representation and apply a lightweight projection along its token axis:
\begin{equation}
    F_t'
    =
    \mathcal{P}_t(F_t^{\top})
    \in
    \mathbb{R}^{C_t\times C_v}.
    \label{eq:text_adapter}
\end{equation}
This maps the textual token dimension to a gating-compatible form while retaining the original embedding dimension \(C_t\). For cooperative cross-gating, \(F_t'\) is broadcast and concatenated with \(v\) along the interaction axis. Specifically, this produces representations of size \(G\times(P+C_t)\times C_v\) and \(P\times(G+C_t)\times C_v\) in the regional and dilated branches, respectively, which are subsequently projected by \(\mathcal{S}(\cdot)\) back to interaction lengths \(P\) and \(G\) for spatial gating. In this way, report-derived semantics directly participate in spatial gate generation for visual features rather than being restricted to a final global conditioning stage.

The two instantiations share the same underlying interaction principle: the mainstream and complementary representations jointly determine spatial modulation through cooperative cross-gating. CoMLP therefore does not require the participating modalities to share the same semantic representation space. Instead, only a lightweight modality-specific adaptation is needed to expose heterogeneous information to a common MLP-based interaction operator, allowing the same fusion primitive to support both inter-image and vision-language information fusion.

\subsection{CoMLP-Based Multi-Source Fusion Architecture}
\label{sec:architecture}
We embed CoMLP into a multi-source fusion architecture to jointly exploit complementary medical images and clinical reports for segmentation. As illustrated in \cref{fig:architecture}, the architecture consists of an inter-image fusion encoder and a vision-language fusion decoder. The former progressively integrates complementary information across imaging modalities, whereas the latter infuses report-derived semantics into multi-scale visual representations.

\subsubsection{Inter-Image Fusion Encoder}
\label{sec:encoder}
Given two input images \(I_{M_1}\) and \(I_{M_2}\), the encoder consists of two modality-specific branches to preserve their individual imaging characteristics while progressively exchanging complementary information. Each branch consists of five successive Hybrid Parallel Visual Learning (HPVL) blocks, with max pooling between adjacent blocks for spatial downsampling. The features generated by the two modality-specific branches at each level are concatenated to construct a five-level multi-modal feature pyramid
\begin{equation}
    F=\{F_1,F_2,F_3,F_4,F_5\},
\end{equation}
whose spatial resolution is progressively reduced from the original resolution to \(1/16\). These multi-scale features are subsequently propagated to the decoder through skip connections.

Each HPVL block performs intra-modal feature learning and inter-image interaction in parallel. The intra-modal pathway uses residual convolutional layers to preserve local modality-specific characteristics, while the interaction pathway first adopts a multi-axis gMLP for intra-feature spatial mixing and then applies CoMLP for explicit inter-image interaction. This parallel design follows the prior principle of jointly preserving intra-modal information and learning inter-modal interactions \cite{meng2023merging}, but replaces attention-based inter-modal interaction with the proposed MLP-based fusion mechanism.

A key characteristic of this encoder is that CoMLP interaction begins from the first, highest-resolution feature level \(F_1\) and continues throughout the subsequent hierarchy. Complementary imaging information can therefore interact before progressive downsampling damages subtle details. As the spatial resolution decreases, successive CoMLP blocks further refine cross-modal dependencies over increasingly abstract representations, resulting in multi-scale inter-image fusion from fine local appearance to broader semantic context.

\subsubsection{Vision-Language Fusion Decoder}
\label{sec:decoder}
The clinical report is encoded using the medical domain-specific language model BiomedBERT \cite{gu2021domain}, yielding a textual representation \(F_t\). The decoder consists of five Text-infused Semantic Guidance (TSG) blocks, with bilinear or trilinear interpolation used for 2D or 3D feature upsampling, respectively.

At each decoding level, a TSG block first combines the upsampled representation from the previous decoder level with the corresponding multi-modal visual feature from the encoder skip connection. Residual convolutional layers are used to consolidate these visual representations. The resulting feature is subsequently treated as the mainstream visual representation and fused with \(F_t\) through CoMLP. Following CoMLP fusion, a Residual Channel Attention (RCA) module is used to recalibrate informative feature channels. The RCA module contains layer normalization, convolutional layers, LeakyReLU activation, and squeeze-and-excitation channel attention \cite{hu2018squeeze}, together with a residual connection. 

Note that report-derived semantic information is not introduced only at the bottleneck or prediction head, but progressively interacts with visual features through successive TSG blocks at multiple spatial scales. After the final TSG block, a convolutional segmentation head followed by a Sigmoid activation generates the predicted segmentation mask.

Through the combination of the encoder and decoder, the same cross-modal interaction primitive is used at two complementary levels. The encoder establishes fine-grained dependencies between heterogeneous imaging modalities, whereas the decoder associates high-level clinical semantics with multi-scale spatial representations. This provides a unified realization of inter-image and vision-language information fusion within a single segmentation architecture.

\subsection{Learning Objective}
\label{sec:objective}
The network is trained end-to-end using the combination of Dice loss and binary cross-entropy loss, following commonly adopted objectives in medical image segmentation \cite{li2023lvit,zhong2023ariadne,lee2023text,guo2024common,zhang2024madapter}. Given the predicted probability \(p_i\) and ground-truth label \(y_i\) for spatial element \(i\), the Dice loss is defined as:
\begin{equation}
    \mathcal{L}_{\mathrm{Dice}}
    =
    1-
    \frac{
    2\sum_i p_i y_i+\epsilon
    }{
    \sum_i p_i+\sum_i y_i+\epsilon
    },
\end{equation}
where \(\epsilon\) is a small constant for numerical stability. The binary cross-entropy loss is defined as:
\begin{equation}
    \mathcal{L}_{\mathrm{CE}}
    =
    -\frac{1}{N}
    \sum_i
    \left[
    y_i\log p_i+
    (1-y_i)\log(1-p_i)
    \right].
\end{equation}
The overall training objective is:
\begin{equation}
    \mathcal{L}
    =
    \mathcal{L}_{\mathrm{Dice}}
    +
    \lambda \mathcal{L}_{\mathrm{CE}},
\end{equation}
where \(\lambda\) controls the relative contribution of the cross-entropy term and is empirically set to \(1\) in our experiments.

\begin{table*}[htb]
\caption{Summary of the comparison methods according to supported information sources and fusion mechanisms.}
\label{tab:fusion_taxonomy}
\centering
\begin{tabular}{@{}p{4cm}
>{\raggedright\arraybackslash}p{6cm}>{\centering\arraybackslash}p{1.5cm}
>{\centering\arraybackslash}p{1.5cm}
>{\centering\arraybackslash}p{1.5cm}
>{\centering\arraybackslash}p{1.5cm}@{}}
\toprule
Method category &
Included methods &
Uni-modal image &
Multi-modal images &
Inter-image fusion &
Text utilization \\
\midrule

General segmentation &
U-Net \cite{ronneberger2015u}, 
U-Net++ \cite{zhou2018unet++}, 
nnU-Net \cite{isensee2021nnu}, 
Swin-Unet \cite{cao2022swin}, 
MLP-Unet \cite{meng2023full} &
\checkmark &
\checkmark &
Concat.&
None \\

Language-guided segmentation &
LViT \cite{li2023lvit}, 
Ariadne's \cite{zhong2023ariadne}, 
MAdapter \cite{zhang2024madapter}, 
ViTexNet \cite{bhardwaj2025vitexnet}, 
FMISeg \cite{yu2025frequency}, 
TeViA \cite{zeng2025harnessing}, 
ConTEXTual Net 3D \cite{huemann2026contextual} &
\checkmark &
\checkmark &
Concat.&
Explicit \\

Multi-modal image segmentation &
SwinCross \cite{li2024swincross}, 
C\textsuperscript{2}MAOT \cite{huang2025c2maot}, 
CIPA \cite{mei2025cross} &
\(\times\) &
\checkmark &
Explicit &
None \\

VLP-based segmentation &
CLIP \cite{radford2021learning}, 
GloRIA \cite{huang2021gloria}, 
BiomedCLIP \cite{zhang2023biomedclip} &
\checkmark &
\(\times\) &
None &
Pretrain\\

\bottomrule
\end{tabular}

\vspace{0.4em}
\begin{minipage}{0.98\textwidth}
\footnotesize
\textit{Note:} ``Explicit'' denotes a dedicated inter-image or vision-language interaction mechanism. ``Pretrain'' denotes using textual information only during representation pretraining. ``Concat.'' denotes channel-wise concatenation of multi-modal images as the input.
\end{minipage}
\end{table*}

\section{Experimental Setup}
\label{sec:experiment}
\subsection{Datasets and Preprocessing}
\label{sec:datasets}
We evaluated CoMLP on five medical image datasets covering 2D/3D images, multiple imaging modalities, clinical reports, and diverse anatomical regions. Among them, the OPC and NPC datasets provide 3D PET/CT images and clinical reports, enabling evaluation of both inter-image and vision-language information fusion. QaTa-COV19, MosMedData+, and Kvasir-SEG provide 2D uni-modal images accompanied by textual descriptions, and are used to independently evaluate the vision-language fusion capability of CoMLP across different imaging modalities and segmentation targets.

\textbf{OPC}: The Oropharynx Cancer (OPC) dataset contains 524 patients acquired from the HECKTOR 2022 challenge \cite{andrearczyk2022overview}. Each case provides paired PET/CT images together with manual segmentation annotations of primary tumors and metastatic lymph nodes. Clinical reports describing lesion location, size, and radioactive tracer uptake were manually written by experienced medical experts based on the PET/CT images, and the medical experts did not have access to the segmentation ground-truth annotations during report preparation. Following \cite{meng2024adaptive}, the dataset was divided into 415/109 cases for training/testing according to clinical centers, and the training set was further randomly split into 400/15 cases for training/validation.

\textbf{NPC}: The Nasopharyngeal Carcinoma (NPC) dataset consists of 866 patients acquired from a clinical study \cite{gu2023multi}. Paired PET/CT images, clinical reports, and segmentation labels of primary tumors and metastatic lymph nodes are provided. The dataset was divided into 632/234 cases for training/testing according to clinical centers, and the training set was further randomly split into 600/32 cases for training/validation.

For OPC and NPC, PET/CT images were resampled into isotropic voxels with a spatial resolution of \(1\times1\times1\) mm\textsuperscript{3} and cropped to \(160\times160\times160\) voxels. PET images were converted into standardized uptake value (SUV) maps according to body mass and subsequently standardized using Z-score normalization. CT images were clipped to the range of [\(-\)1024, 1024] HU and then normalized to [\(-\)1, 1].

\textbf{QaTa-COV19}: QaTa-COV19 \cite{degerli2022osegnet} contains 9258 chest X-ray images with COVID-19 lesion annotations. Li et al. \cite{li2023lvit} further provided textual descriptions detailing bilateral pulmonary infection, the number of affected regions, and their spatial locations. We followed the official split of 5716, 1429, and 2113 images for training, validation, and testing.

\textbf{MosMedData+}: MosMedData+ \cite{morozov2020mosmeddata} contains 2729 CT slices depicting pulmonary infections with segmentation annotations and corresponding textual descriptions. We followed the official split of 2183, 273, and 273 images for training, validation, and testing.

\textbf{Kvasir-SEG}: Kvasir-SEG \cite{jha2019kvasir} contains 1000 colonoscopy images with segmentation annotations of gastrointestinal polyps. We adopted the established textual annotations \cite{poudel2023exploring} describing polyp size, number, color, and location for language-guided segmentation. Following \cite{poudel2023exploring}, the dataset was divided with a ratio of 8:1:1 for training, validation, and testing.

For the three 2D datasets, the images were resized to \(224\times224\) and normalized to the range of [0, 1].

\subsection{Implementation Details}
\label{sec:implementation}
Our implementation is based on PyTorch and all experiments were conducted using an NVIDIA A100 GPU with 40 GB memory. Our method was optimized using Adam with a batch size of 32 for 2D images and 2 for 3D images. The models were trained for 20,000 iterations. The initial learning rate was set to \(1\times10^{-4}\) and successively reduced to \(5\times10^{-5}\), \(1\times10^{-5}\), and \(1\times10^{-6}\) after 4,000, 8,000, and 12,000 iterations, respectively. Random translation, scaling, shearing, and rotation were performed online for data augmentation. Validation was conducted every 200 training iterations.

The initial visual embedding dimension was set to 16, while the textual embedding dimension produced by BiomedBERT \cite{gu2021domain} was 768. According to the hyperparameter analysis in \cref{sec:Hyperparameter}, the region size \(R\) was set to 12 for the 2D datasets and 8 for the 3D datasets.

For OPC and NPC, our CoMLP-based multi-source architecture was used to jointly process PET/CT images and clinical reports. For QaTa-COV19, MosMedData+, and Kvasir-SEG, which contain only a single imaging modality, the inter-image fusion encoder was replaced by a uni-modal encoder composed of hybrid CNN-MLP blocks with parallel convolution and multi-axis gMLP \cite{meng2023full}, while the vision-language fusion decoder remained unchanged. In the experiments designed to isolate inter-image fusion on OPC and NPC, clinical reports were excluded and the vision-language decoder was replaced by an Attention U-Net decoder \cite{schlemper2019attention}. These configurations allow the inter-image and vision-language instantiations of CoMLP to be evaluated independently.

\begin{table*}[htb]
  \caption{Quantitative comparison on OPC and NPC with multi-modal images and clinical reports.}
  \label{tab:overall_3d}
  \centering
  \begin{tabular}{@{}>{\centering\arraybackslash}p{6cm}>{\centering\arraybackslash}p{1.6cm}
  >{\centering\arraybackslash}p{1.6cm}
  >{\centering\arraybackslash}p{1.6cm}
  >{\centering\arraybackslash}p{1.6cm}
  >{\centering\arraybackslash}p{1.6cm}
  >{\centering\arraybackslash}p{1.6cm}@{}}
    \toprule
    \multirow{2}{*}{Method}&
    \multicolumn{2}{c}{OPC}&
    \multicolumn{2}{c}{NPC}&
    \multicolumn{2}{c}{Average}\\
    & Dice (\%) & mIoU (\%) & Dice (\%) & mIoU (\%) & Dice (\%) & mIoU (\%) \\
    \midrule

    \multicolumn{7}{l}{{\it General segmentation (concatenated multi-modal images; no report)}}\\
    U-Net \cite{ronneberger2015u}
    &74.29&60.95&75.37&61.73&74.83&61.34\\
    U-Net++ \cite{zhou2018unet++}
    &75.18&61.66&75.88&62.29&75.53&61.98\\
    nnU-Net \cite{isensee2021nnu}
    &76.23&62.98&76.89&63.22&76.56&63.10\\
    Swin-Unet \cite{cao2022swin}
    &75.95&62.80&76.53&62.84&76.24&62.82\\
    MLP-Unet \cite{meng2023full}
    &77.76&63.43&78.61&64.55&78.19&63.99\\

    \midrule
    \multicolumn{7}{l}{{\it Multi-modal image segmentation (explicit inter-image fusion; no report)}}\\
    SwinCross \cite{li2024swincross}
    &77.62&63.68&78.84&64.76&78.23&64.22\\
    C\textsuperscript{2}MAOT \cite{huang2025c2maot}
    &78.53&64.71&79.57&66.03&79.05&65.37\\
    CIPA \cite{mei2025cross}
    &78.66&64.50&79.23&65.73&78.95&65.16\\
    CoMLP (w/o report)
    &79.78&65.79&81.24&68.15&80.51&66.97\\

    \midrule
    \multicolumn{7}{l}{{\it Language-guided segmentation (concatenated multi-modal images + explicit vision-language fusion)}}\\
    LViT \cite{li2023lvit}
    &74.77&61.24&75.85&62.24&75.31&61.74\\
    Ariadne's \cite{zhong2023ariadne}
    &75.14&62.11&76.51&62.54&75.83&62.33\\
    MAdapter \cite{zhang2024madapter}
    &76.77&63.11&77.50&63.56&77.14&63.34\\
    ViTexNet \cite{bhardwaj2025vitexnet}
    &77.90&63.72&79.43&65.54&78.67&64.63\\
    FMISeg \cite{yu2025frequency}
    &77.59&63.89&78.67&64.85&78.13&64.37\\
    TeViA \cite{zeng2025harnessing}
    &77.14&63.22&78.85&64.38&78.00&63.80\\
    ConTEXTual Net 3D \cite{huemann2026contextual}
    &77.25&63.86&78.58&65.58&77.92&64.72\\

    \midrule
    \multicolumn{7}{l}{{\it Explicit inter-image + vision-language fusion}}\\
    CoMLP (Ours)
    &{\bf 81.24$^{*}$}&{\bf 67.12$^{*}$}&{\bf 82.95$^{*}$}&{\bf 69.86$^{*}$}
    &{\bf 82.10}&{\bf 68.49}\\
    \bottomrule
  \end{tabular}

\vspace{0.4em}
\begin{minipage}{0.98\textwidth}
\footnotesize
\textit{Note:} \textbf{Bold}: Best result in each column. $^{*}$: Better than all comparison methods with statistical significance ($p<0.05$, after Holm--Bonferroni correction).
\end{minipage}
\end{table*}

\begin{table*}[htb]
  \caption{Quantitative comparison on QaTa-COV19, MosMedData+, and Kvasir-SEG with uni-modal images and textual descriptions.}
  \label{tab:overall_2d}
  \centering
  \begin{tabular}{@{}>{\centering\arraybackslash}p{3.5cm}>{\centering\arraybackslash}p{1.35cm}
  >{\centering\arraybackslash}p{1.45cm}
  >{\centering\arraybackslash}p{1.35cm}
  >{\centering\arraybackslash}p{1.45cm}
  >{\centering\arraybackslash}p{1.35cm}
  >{\centering\arraybackslash}p{1.45cm}
  >{\centering\arraybackslash}p{1.35cm}
  >{\centering\arraybackslash}p{1.45cm}@{}}
    \toprule
    \multirow{2}{*}{Method}&
    \multicolumn{2}{c}{QaTa-COV19}&
    \multicolumn{2}{c}{MosMedData+}&
    \multicolumn{2}{c}{Kvasir-SEG}&
    \multicolumn{2}{c}{Average}\\
    &Dice (\%)&mIoU (\%)&Dice (\%)&mIoU (\%)&Dice (\%)&mIoU (\%)&Dice (\%)&mIoU (\%)\\
    \midrule

    \multicolumn{9}{l}{{\it General segmentation (no text)}}\\
    U-Net \cite{ronneberger2015u}
    &82.99&70.92&64.60&50.73&82.09&69.62&76.56&63.76\\
    U-Net++ \cite{zhou2018unet++}
    &83.69&71.96&71.75&58.39&82.06&69.58&79.17&66.64\\
    nnU-Net \cite{isensee2021nnu}
    &80.42&70.81&72.59&60.36&/&/&/&/\\
    Swin-Unet \cite{cao2022swin}
    &83.60&72.40&66.90&53.10&82.41&72.38&77.64&65.96\\
    MLP-Unet \cite{meng2023full}
    &85.22&77.48&75.87&62.86&86.27&77.46&82.45&72.60\\

    \midrule
    \multicolumn{9}{l}{{\it VLP-based segmentation (text used for pretraining)}}\\
    CLIP \cite{radford2021learning}
    &79.81&70.66&71.97&59.64&/&/&/&/\\
    GloRIA \cite{huang2021gloria}
    &79.94&70.68&72.42&60.18&/&/&/&/\\
    BiomedCLIP \cite{zhang2023biomedclip}
    &87.88&78.38&66.51&50.52&85.66&77.91&80.02&68.94\\

    \midrule
    \multicolumn{9}{l}{{\it Language-guided segmentation (explicit vision-language fusion)}}\\
    LViT \cite{li2023lvit}
    &83.66&75.11&74.57&61.33&87.03&77.04&81.75&71.16\\
    Ariadne's \cite{zhong2023ariadne}
    &89.78&81.45&77.75&63.60&89.32&80.70&85.62&75.25\\
    MAdapter \cite{zhang2024madapter}
    &90.22&82.16&78.62&64.78&91.58&84.47&86.81&77.14\\
    ViTexNet \cite{bhardwaj2025vitexnet}
    &90.76&83.25&78.19&64.04&91.44&84.62&86.90&77.30\\
    FMISeg \cite{yu2025frequency}
    &91.21&83.84&79.30&65.71&91.12&83.52&87.21&77.69\\
    TeViA \cite{zeng2025harnessing}
    &91.06&83.60&78.49&64.59&90.92&83.77&86.82&77.32\\
    CoMLP (Ours)
    &{\bf 91.55}&{\bf 84.37}
    &{\bf 80.31}&{\bf 67.19}
    &{\bf 92.85}&{\bf 87.06}
    &{\bf 88.24}&{\bf 79.54}\\
    \bottomrule
  \end{tabular}

\vspace{0.4em}
\begin{minipage}{0.98\textwidth}
\footnotesize
\textit{Note:} \textbf{Bold}: Best result in each column. CLIP and GLoRIA were not evaluated on Kvasir-SEG as they were not pretrained on this domain.
\end{minipage}
\end{table*}

\subsection{Comparison Methods}
\label{sec:ComMethod}
We compared CoMLP with representative medical image segmentation methods according to their supported information sources and corresponding fusion mechanisms. As summarized in \cref{tab:fusion_taxonomy}, the comparison methods can be grouped into four categories: (i) General segmentation methods can be directly applied to either uni-modal images or multi-modal images, with the latter handled through channel-wise input concatenation. (ii) Language-guided segmentation methods explicitly incorporate textual information during segmentation and can be applied to uni-modal images or adapted to multi-modal images by concatenating the multi-modal images as the visual input. (iii) Multi-modal image segmentation methods are specifically designed for multiple imaging modalities and explicitly model inter-image information interaction. (iv) Vision-language pretraining (VLP) methods considered in this study operate on uni-modal images and exploit image-text correspondence during representation pretraining, but do not explicitly use the text during segmentation inference. 

The comparison methods included in each category are reported in \cref{tab:fusion_taxonomy}. For OPC and NPC, general and language-guided segmentation methods receive concatenated PET/CT as the visual input, whereas the multi-modal image segmentation methods and CoMLP explicitly model interactions between the two imaging modalities. Accordingly, comparisons on OPC and NPC evaluate methods under multi-modal imaging inputs, with or without clinical reports depending on the method category. For methods originally developed for 2D images, we extended their visual operations to 3D counterparts, while preserving their original network architecture, vision-language interaction mechanisms, and other method-specific designs. For all re-implemented comparison methods, the training schedules were re-tuned on the validation sets. Comparisons on QaTa-COV19, MosMedData+, and Kvasir-SEG are conducted under the uni-modal setting, where general segmentation, VLP-based, and language-guided segmentation methods are applicable. The comparison results on the three public 2D benchmarks were taken from the corresponding original publications if available. 

\begin{figure*}[htb]
  \centering
  \includegraphics[height=12.5cm]{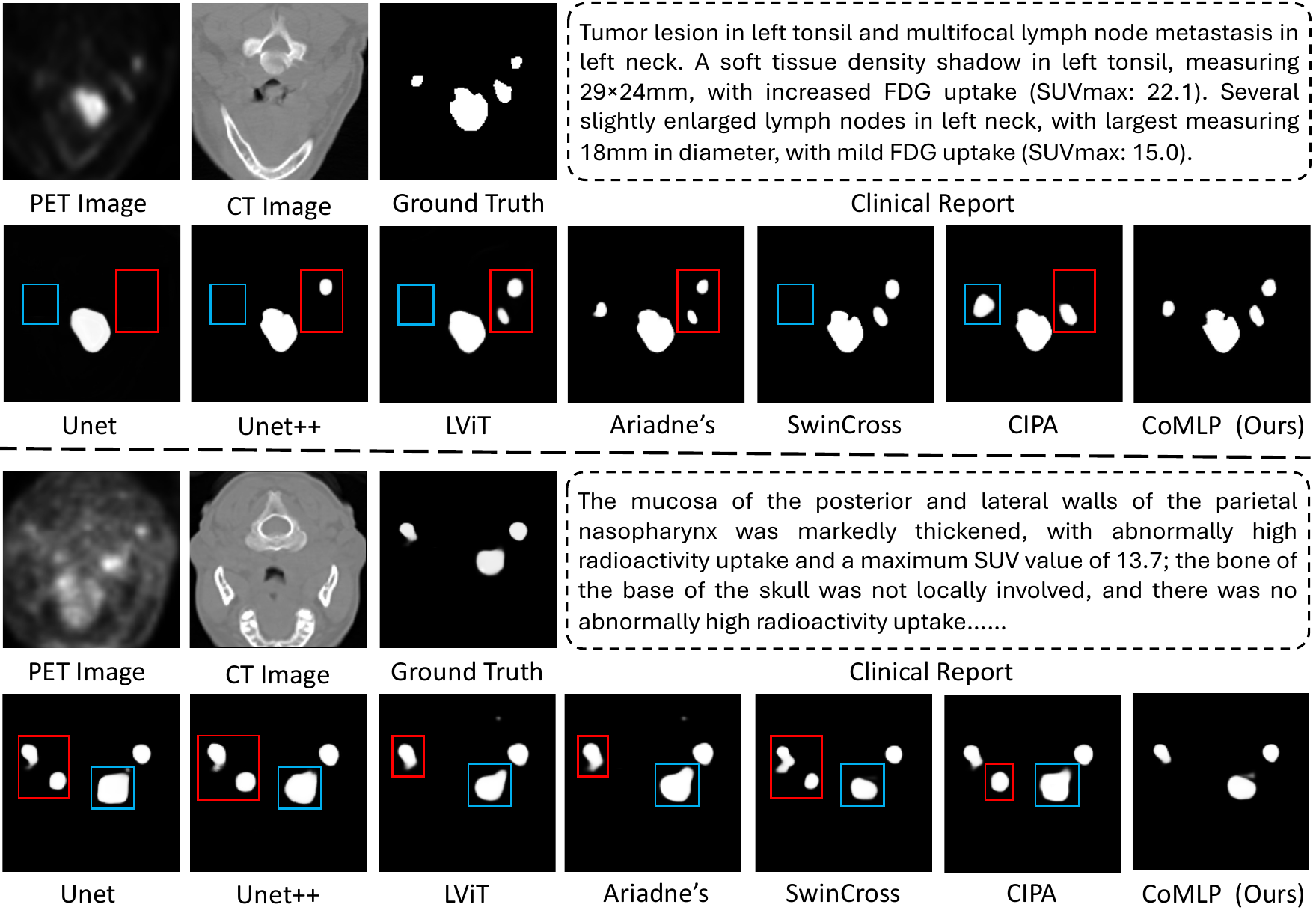}
  \caption{Qualitative comparison on OPC (top) and NPC (bottom). Representative differences between the predicted segmentation masks and ground-truth annotations  are highlighted in colored boxes.}
  \label{fig:fig3}
\end{figure*}

\subsection{Experimental Designs}
\label{sec:expdesign}
Our experiments were designed to evaluate CoMLP from multiple complementary perspectives. First, we compared the proposed CoMLP-based method with existing segmentation methods on all five datasets to assess its overall effectiveness across different combinations of imaging modalities and textual information. Second, we independently investigated the two applications of CoMLP, namely inter-image fusion and vision-language fusion, through controlled ablation studies. These experiments examine the importance of high-resolution interaction, the effectiveness of cooperative cross-gating relative to conventional cross-gating and cross-attention, and the influence of preserving heterogeneous visual and textual representations. Then, we further analyzed the underlying properties of CoMLP beyond individual fusion settings. We evaluated whether the same interaction primitive consistently benefits both inter-image and vision-language fusion, investigated the relationship between computational efficiency and the practically achievable fusion granularity, and examined the complementary roles of the regional and dilated interaction branches. Finally, a hyperparameter analysis was conducted to explore the influence of the region size \(R\). Detailed configurations are introduced together with the corresponding results in the following sections. 

The Dice Similarity Coefficient (Dice) and mean Intersection over Union (mIoU) were adopted, following existing multi-modal and language-guided medical image segmentation studies \cite{li2023lvit,zhong2023ariadne,lee2023text,guo2024common,zhang2024madapter}. Both metrics quantify the spatial overlap between predicted and ground-truth segmentations, with higher values indicating better performance. For OPC and NPC, statistical significance was assessed using two-sided Wilcoxon signed-rank tests on paired per-case results, with Holm-Bonferroni correction for multiple comparisons across methods. Paired statistical testing was not performed on the three public 2D benchmarks because only published aggregate results were available for the comparison methods.

\begin{table*}[htb]
  \caption{Analysis of inter-image fusion on OPC and NPC without using clinical reports.}
  \label{tab:interimage}
  \centering
  \begin{tabular}{@{}>{\centering\arraybackslash}p{6.5cm}>{\centering\arraybackslash}p{1.4cm}
  >{\centering\arraybackslash}p{1.5cm}
  >{\centering\arraybackslash}p{1.4cm}
  >{\centering\arraybackslash}p{1.5cm}
  >{\centering\arraybackslash}p{1.4cm}
  >{\centering\arraybackslash}p{1.5cm}@{}}
    \toprule
    \multirow{2}{*}{Configuration}&
    \multicolumn{2}{c}{OPC}&
    \multicolumn{2}{c}{NPC}&
    \multicolumn{2}{c}{Average}\\
    &Dice (\%)&mIoU (\%)&Dice (\%)&mIoU (\%)&Dice (\%)&mIoU (\%)\\
    \midrule
    CoMLP (\(F_1\)-\(F_5\))&{\bf 79.78}&{\bf 65.79}&{\bf 81.24}&{\bf 68.15}&{\bf 80.51}&{\bf 66.97}\\
    CoMLP (\(F_2\)-\(F_5\))&77.80&63.71&79.36&65.85&78.58&64.78\\
    CoMLP (\(F_3\)-\(F_5\))&76.32&62.55&77.97&64.21&77.15&63.38\\
    CoMLP (\(F_4\)-\(F_5\))&75.35&61.81&76.84&63.18&76.10&62.50\\
    CoMLP (\(F_5\))
    &74.83&61.37&76.25&62.34&75.54&61.86\\
    Remove all CoMLP
    &74.56&61.05&75.83&61.99&75.20&61.52\\
    \midrule
    MA-CG (\(F_1\)-\(F_5\))& 78.89& 64.63& 80.45& 66.98& 79.67&65.81\\
    MA-CG (\(F_2\)-\(F_5\))& 77.25& 63.07& 78.43& 64.35& 77.84&63.71\\
    MA-CA (\(F_2\)-\(F_5\))&77.43&63.44&78.58&64.68&78.01&64.06\\
    Global-CA (\(F_5\))& 75.24& 61.72& 76.38& 62.77& 75.81&62.25\\
    \bottomrule
  \end{tabular}

\vspace{0.4em}
\begin{minipage}{0.98\textwidth}
\footnotesize
\textit{Note:} ``MA-CG/CA'': Replace cooperative cross-gating with conventional cross-gating \cite{tu2022maxim}/cross-attention while retaining the regional/dilated design. ``Global-CA'': Global cross-attention between complete feature maps. \textbf{Bold}: Best result in each column.
\end{minipage}
\end{table*}

\begin{table*}[htb]
  \caption{Analysis of vision-language fusion on QaTa-COV19, MosMedData+, and Kvasir-SEG.}
  \label{tab:visionlanguage}
  \centering
  \begin{tabular}{@{}>{\centering\arraybackslash}p{3cm}>{\centering\arraybackslash}p{1.35cm}
  >{\centering\arraybackslash}p{1.45cm}
  >{\centering\arraybackslash}p{1.35cm}
  >{\centering\arraybackslash}p{1.45cm}
  >{\centering\arraybackslash}p{1.35cm}
  >{\centering\arraybackslash}p{1.45cm}
  >{\centering\arraybackslash}p{1.35cm}
  >{\centering\arraybackslash}p{1.45cm}@{}}
    \toprule
    \multirow{2}{*}{Configuration}&
    \multicolumn{2}{c}{QaTa-COV19}&
    \multicolumn{2}{c}{MosMedData+}&
    \multicolumn{2}{c}{Kvasir-SEG}&
    \multicolumn{2}{c}{Average}\\
    &Dice (\%)&mIoU (\%)&Dice (\%)&mIoU (\%)&Dice (\%)&mIoU (\%)&Dice (\%)&mIoU (\%)\\
    \midrule
    CoMLP (Text dim. 768)
    &{\bf 91.55}&{\bf 84.37}&{\bf 80.31}&{\bf 67.19}
    &{\bf 92.85}&{\bf 87.06}&{\bf 88.24}&{\bf 79.54}\\
    CoMLP (Text dim. 384)
    &91.38&84.02&80.15&66.98&92.37&86.29&87.97&79.10\\
    CoMLP (Text dim. 192)
    &90.79&83.54&79.88&66.35&91.53&84.98&87.40&78.29\\
    CoMLP (Text dim. 96)
    &90.32&83.12&79.43&65.69&90.89&84.04&86.88&77.62\\
    CoMLP (Text dim. 48)
    &89.77&82.05&78.76&64.88&90.17&83.35&86.23&76.76\\
    CoMLP (Text dim. 24)
    &89.24&81.27&77.99&63.93&89.53&82.19&85.59&75.80\\
    \midrule
    MA-CG& 90.27& 82.55& 78.74& 64.79& 90.53& 83.86& 86.51&77.07\\
    MA-CA& 90.43& 82.64& 78.81& 65.06& 90.82& 84.11& 86.69&77.27\\
    Global-CA&89.85&81.98&78.26&64.38&89.72&82.36&85.94&76.24\\
    \bottomrule
  \end{tabular}

\vspace{0.4em}
\begin{minipage}{0.98\textwidth}
\footnotesize
\textit{Note:} ``MA-CG/CA'': Replace cooperative cross-gating with conventional cross-gating \cite{tu2022maxim}/cross-attention while retaining the regional/dilated design. ``Global-CA'': Global cross-attention between complete feature maps. ``Text dim.'': Textual embedding dimension. \textbf{Bold}: Best result in each column.
\end{minipage}
\end{table*}

\section{Results}
\label{sec:results}
\subsection{Overall Comparison with Existing Methods}
\label{sec:overall}
\cref{tab:overall_3d} compares different strategies for exploiting multi-modal images and clinical reports on OPC and NPC. General segmentation methods use both imaging modalities through concatenation but do not explicitly model their cross-modal dependencies. Explicit inter-image fusion consistently improves over this early-fusion strategy, as demonstrated by SwinCross, C\textsuperscript{2}MAOT, CIPA, and the no-report variant of CoMLP. CoMLP without reports still outperforms the existing multi-modal image segmentation methods, indicating that explicitly modeling fine-grained inter-image dependencies is already beneficial even without textual information.

Language-guided segmentation methods additionally exploit clinical reports but generally concatenate multi-modal images before vision-language interaction. Their improvements over general segmentation methods confirm the complementary value of textual semantics, whereas their overall performance remains constrained by the lack of explicit inter-image interaction. Our CoMLP jointly performs explicit inter-image and vision-language fusion and thereby achieves the best performance on both datasets. 

\cref{tab:overall_2d} further evaluates the vision-language fusion capability of CoMLP independently from inter-image fusion. Language-guided segmentation methods generally outperform general segmentation and VLP-based methods, demonstrating that directly incorporating textual information during segmentation is more effective than using image-text correspondence only during representation pretraining. CoMLP consistently achieves the best performance across all three datasets, indicating that the proposed mechanism is not restricted to particular imaging modalities and anatomical regions.

\subsection{Qualitative Comparison}
\label{sec:qualitative}
As shown in \cref{fig:fig3}, CoMLP produces segmentation masks that are visually more consistent with the ground-truth annotations than the comparison methods. Two representative improvements can be observed. First, CoMLP more reliably recovers small or spatially localized lesions that are missed by the comparison methods, which is consistent with its ability to perform fine-grained cross-modal interaction at high-resolution feature maps. Second, CoMLP reduces false-positive responses in visually confusing regions. This observation is consistent with the use of broader cross-modal textual contexts to distinguish target lesions from structures with similar appearance. These qualitative observations complement the quantitative improvements.

\subsection{Inter-Image Fusion with CoMLP}
\label{sec:interimage_results}
To isolate inter-image fusion, we conducted experiments on OPC and NPC without using clinical reports. We investigated how the highest spatial resolution of explicit inter-image interaction affects segmentation performance. Starting from the full CoMLP configuration, the inter-image interaction was progressively removed from \(F_1\) to \(F_5\), where \(F_1\) denotes the highest-resolution feature level.

As shown in \cref{tab:interimage}, performance consistently decreases as the highest interaction resolution is reduced. Removing only \(F_1\) results in the largest single degradation among adjacent configurations. Further removing interaction from progressively lower-resolution levels produces additional performance degradation. This confirms that inter-image fusion benefits strongly from interaction at fine spatial granularity, particularly before subtle information is damaged by downsampling.

We compare CoMLP with cross-attention and conventional cross-gating variants to investigate the effects of the interaction operator and practically achievable fusion granularity. Global-CA performs unrestricted cross-attention between complete feature maps. Due to its rapidly increasing memory consumption, Global-CA is applied at the lowest spatial resolution \(F_5\) under our GPU memory budget (see \cref{sec:efficiency}). MA-CA retains the multi-axis design and can extend interaction from \(F_5\) to \(F_2\) (see \cref{sec:efficiency}), but its performance remains below CoMLP at the matched highest resolution \(F_2\). We further introduce MA-CG, which retains the same multi-axis design but replaces cooperative cross-gating with conventional cross-gating. At \(F_2\)-\(F_5\), MA-CG outperforms Global-CA but remains below MA-CA. Extending MA-CG to the full-resolution \(F_1\) further improves the performance and exceeds MA-CA operating from \(F_2\), highlighting the importance of finer interaction granularity. Nevertheless, CoMLP remains superior to MA-CG under the same \(F_1\)-\(F_5\) configuration. These results demonstrate that fine-grained interaction at high spatial resolutions is a key factor in inter-image fusion, while cooperative cross-gating provides further improvement under matched interaction granularity.

\subsection{Vision-Language Fusion with CoMLP}
\label{sec:vl_results}
We further isolate vision-language fusion on QaTa-COV19, MosMedData+, and Kvasir-SEG, where only one imaging modality is involved. Since the textual sequence is relatively short, vision-language cross-attention does not suffer from the same severe computational bottleneck as inter-image attention. We hence focus on the interaction mechanism and the preservation of heterogeneous textual representations in \cref{tab:visionlanguage}.

We first compare CoMLP with Global-CA, which directly establishes cross-attention between visual and textual features but yields lower performance than CoMLP. We further compare two variants under the same regional/dilated multi-axis design. MA-CG replaces cooperative cross-gating with conventional cross-gating, improving over Global-CA but remaining below MA-CA. In conventional cross-gating, the modulation of the mainstream feature is generated solely from the complementary representation, such that gate construction does not explicitly depend on the current mainstream information. This cross-conditioned modulation may therefore be less effective for heterogeneous vision-language interaction, where textual semantics need to be selectively associated with different visual contents. By jointly incorporating the mainstream and cross-modal information, CoMLP enables content-adaptive gate generation and remains superior to both variants.

As Global-CA, MA-CA, and MA-CG are all computationally feasible in vision-language fusion, the benefit of CoMLP in this setting is not attributed to avoiding the computational burden of attention. The textual embedding analysis instead provides another perspective on representation preservation. CoMLP allows the original high-dimensional textual representation to be retained before gate adaptation, such that the full-dimensional textual representation remains directly involved in cooperative gate generation. Progressively reducing the textual embedding dimension from 768 to 24 produces a consistent performance decrease, suggesting that retaining richer textual representations before cross-modal interaction is beneficial for effective semantic guidance. These results highlight the practical benefit of preserving high-dimensional textual information within the cooperative gating process.

\subsection{Unified Interaction across Heterogeneous Sources}
\label{sec:unified_results}
We further evaluate whether CoMLP can consistently benefit both inter-image and vision-language fusion within the same OPC/NPC segmentation setting. As shown in \cref{tab:unified}, using MA-CA for both interactions yields an average Dice/mIoU of 79.55/65.72\%. Replacing MA-CA with CoMLP only for inter-image fusion improves the performance to 81.34/67.55\%, while replacing it only for vision-language fusion increases the results to 80.27/66.14\%. These results demonstrate that CoMLP provides complementary benefits in both heterogeneous fusion processes.

When CoMLP is applied to both inter-image and vision-language fusion, the performance further reaches 82.10\% Dice and 68.49\% mIoU, outperforming the dual MA-CA configuration by 2.55 and 2.77 percentage points, respectively. Note that MA-CA starts inter-image interaction from \(F_2\) due to its memory demand, whereas CoMLP supports interaction beginning from the full-resolution \(F_1\). The results reflect both the effectiveness of cooperative cross-gating and the finer inter-image fusion granularity practically enabled by CoMLP. Overall, the consistent gains across both stages support CoMLP as a common interaction primitive for heterogeneous inter-image and vision-language information fusion.

\begin{table}[tb]
\caption{Unified interaction across heterogeneous information sources on OPC and NPC.}
\label{tab:unified}
\centering
\begin{tabular}{@{}>{\centering\arraybackslash}p{2cm}>{\centering\arraybackslash}p{2cm}cc@{}}
\toprule
Inter-image&
Vision-language&
Avg. Dice (\%) &
Avg. mIoU (\%) \\
\midrule
MA-CA & MA-CA & 79.55& 65.72\\
CoMLP & MA-CA & 81.34& 67.55\\
MA-CA & CoMLP & 80.27& 66.14\\
CoMLP& CoMLP & \textbf{82.10} & \textbf{68.49} \\
\bottomrule
\end{tabular}

\vspace{0.4em}
\begin{minipage}{0.48\textwidth}
\footnotesize
\textit{Note:} ``MA-CA'': Replace cooperative cross-gating with cross-attention while retaining the same multi-axis design. \textbf{Bold}: Best result in each column.
\end{minipage}
\end{table}

\begin{table}[tb]
\caption{Ablation of the regional and dilated interaction branches.}
\label{tab:branches}
\centering
\begin{tabular}{@{}>{\raggedright\arraybackslash}p{2.4cm}>{\centering\arraybackslash}p{1.15cm}>{\centering\arraybackslash}p{1.2cm}>{\centering\arraybackslash}p{1.15cm}>{\centering\arraybackslash}p{1.2cm}@{}}
\toprule
Configuration &
\multicolumn{2}{c}{Inter-image} &
\multicolumn{2}{c}{Vision-language}\\
& Dice (\%)& mIoU (\%)& Dice (\%)& mIoU (\%)\\
\midrule
Full CoMLP
&\textbf{80.51}&\textbf{66.97}&\textbf{88.24}&\textbf{79.54}\\
w/o Dilated branch
&79.57&65.89&84.46&74.54\\
w/o Regional branch
&78.66&64.90&85.95&76.03\\
\bottomrule
\end{tabular}

\vspace{0.4em}
\begin{minipage}{0.48\textwidth}
\footnotesize
\textit{Note:} ``Inter-image" denotes results averaged across OPC and NPC. ``Vision-language" denotes results averaged across QaTa-COV19, MosMedData+, and Kvasir-SEG. \textbf{Bold}: Best result in each column.
\end{minipage}
\end{table}

\begin{table*}[htb]
\caption{Computational efficiency analysis of inter-image fusion on OPC/NPC without using clinical reports.}
\label{tab:efficiency}
\centering
\begin{tabular}{@{}>{\centering\arraybackslash}p{2.2cm}ccccc>{\centering\arraybackslash}p{1.3cm}>{\centering\arraybackslash}p{1.3cm}>{\centering\arraybackslash}p{1.2cm}>{\centering\arraybackslash}p{1.2cm}>{\centering\arraybackslash}p{1.5cm}>{\centering\arraybackslash}p{1.5cm}@{}}
\toprule
Operator&
\(F_5\) &
\(F_4\) &
\(F_3\) &
\(F_2\) &
\(F_1\) &
Training Memory&Inference Memory& Parameter Number&FLOPs per Case&
Avg. Dice (\%)&
Avg. mIoU (\%)\\
\midrule
Global-CA&
\checkmark &
OOM &
OOM&
OOM&
OOM&
24GB& 5.0GB& 24.7M&0.86T&
75.81&
62.25\\

MA-CA &
\checkmark &
\checkmark &
\checkmark &
\checkmark &
OOM &
38GB& 7.5GB& 26.5M&0.98T&
78.01 &
64.06 \\

CoMLP &
\checkmark &
\checkmark &
\checkmark &
\checkmark &
\(\times\)&
24GB& 5.3GB& 28.5M&0.90T&
78.58&
64.78\\

CoMLP &
\checkmark &
\checkmark &
\checkmark &
\checkmark &
\checkmark &
28GB& 5.9GB& 30.6M&1.14T&
80.51 &
66.97 \\
\bottomrule
\end{tabular}

\vspace{0.4em}
\begin{minipage}{0.98\textwidth}
\footnotesize
\textit{Note:} ``OOM" denotes out of memory during training on a 40GB NVIDIA A100. \(F_n\) denotes inter-image interaction occurring in this resolution level.
\end{minipage}
\end{table*}

\begin{figure*}[htb]
  \centering
  \includegraphics[height=5cm]{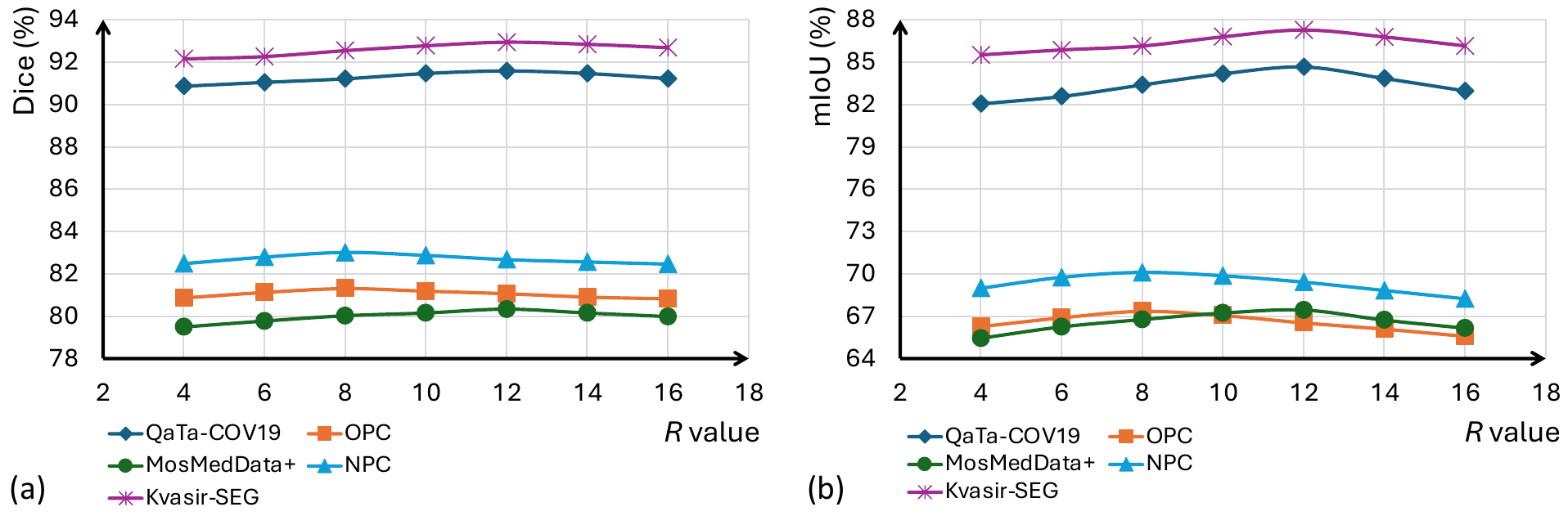}
  \caption{Hyperparameter analysis of the region size \(R\) in CoMLP. (a) and (b) present Dice and mIoU results under different \(R\) values, respectively.}
  \label{fig:fig4}
\end{figure*}

\subsection{Analysis of Regional and Dilated Interactions}
\label{sec:branch_analysis}
We separately examine the regional and dilated interaction branches to determine whether local and long-range cross-modal dependency modeling provide complementary benefits for inter-image and vision-language fusion.

As shown in \cref{tab:branches}, removing either branch consistently degrades performance, confirming that regional and dilated interactions provide complementary information in both fusion settings. For inter-image fusion, removing the regional branch produces a larger decrease than removing the dilated branch. This suggests that fine local correspondence is particularly important when integrating spatially aligned imaging modalities such as PET and CT, where complementary anatomical and metabolic details need to be associated at local spatial locations. For vision-language fusion, removing the dilated branch causes a larger degradation. This indicates that report-derived semantics benefit strongly from broader contextual interaction across spatially distributed visual regions. Nevertheless, retaining both branches consistently provides the best performance, supporting the design of CoMLP as a complementary local-global interaction mechanism rather than relying exclusively on either type of dependency.

\subsection{Computational Efficiency and Fusion Granularity}
\label{sec:efficiency}
We further investigate how computational efficiency affects the practically achievable granularity of inter-image fusion. As shown in \cref{tab:efficiency}, Global-CA performs unrestricted cross-attention over complete feature maps and can only operate at the lowest resolution \(F_5\) within the 40GB GPU budget. MA-CA substantially improves scalability through the regional/dilated multi-axis design and extends interaction to \(F_2\), but already requires 38 GB of training memory and becomes infeasible at the full-resolution \(F_1\).

A direct comparison at the matched fusion resolution \(F_2\)-\(F_5\) clearly demonstrates the efficiency of CoMLP. Compared with MA-CA, CoMLP reduces the training/inference memory from 38/7.5 GB to 24/5.3 GB and the computational cost from 0.98 to 0.90 TFLOPs, while improving the average Dice/mIoU from 78.01/64.06\% to 78.58/64.78\%. Although CoMLP has slightly more parameters (28.5 M vs. 26.5 M), it avoids memory-intensive pairwise attention matrices and thus provides substantially better memory efficiency.

This improved memory efficiency enables CoMLP to extend inter-image interaction to the full-resolution \(F_1\). The resulting configuration increases computation to 1.14 TFLOPs and parameters to 30.6 M, but still requires only 28/5.9 GB of training/inference memory, remaining well within the same hardware budget. Importantly, the finer interaction granularity further improves the average Dice/mIoU to 80.51/66.97\%. These results demonstrate that the main efficiency advantage of CoMLP lies not in minimizing computational cost, but in enabling finer-grained cross-modal interaction at higher spatial resolutions that are practically inaccessible to fusion operators based on cross-attention.

\subsection{Hyperparameter Analysis}
\label{sec:Hyperparameter}
\cref{fig:fig4} presents the validation performance under different region sizes \(R\) on all five datasets. Both Dice and mIoU vary smoothly as \(R\) changes from 4 to 16, indicating that CoMLP is generally robust to this hyperparameter. The best validation results are obtained with \(R=8\) on the two 3D datasets, OPC and NPC, whereas \(R=12\) performs best on the three 2D datasets, QaTa-COV19, MosMedData+, and Kvasir-SEG.

The region size determines the complementary operating ranges of the two interaction branches. When \(R\) is too small, the regional branch covers only a limited local neighborhood and may fail to capture sufficient regional context. In contrast, an excessively large \(R\) increases the spatial interval between locations interacting in the dilated branch, making long-range modeling increasingly sparse. An intermediate region size therefore provides a better balance between fine local interaction and broader contextual dependency modeling.

\section{Discussion and Limitations}
\label{sec:discussion}
Our study identifies that the effectiveness of cross-modal fusion depends not only on the interaction operator, but also on the spatial granularity of information interaction. For inter-image fusion, progressively removing interaction from high-resolution levels consistently degrades segmentation, with the largest decrease occurring after removing the full-resolution interaction. The efficiency analysis further shows that this fusion granularity is closely coupled with memory scalability. At the matched \(F_2\)-\(F_5\) resolution, CoMLP achieves comparable computational complexity to MA-CA while requiring substantially less training and inference memory; this efficiency further enables interaction at \(F_1\), where MA-CA becomes infeasible, and produces additional performance gains. Therefore, the main computational benefit of CoMLP is not simply lower FLOPs or fewer parameters, but making finer-grained cross-modal interaction practically accessible before subtle visual information is degraded by downsampling.

The experiments also demonstrate that CoMLP can serve as a common interaction primitive for substantially different heterogeneous information sources. Replacing MA-CA with CoMLP independently for either inter-image or vision-language fusion improves segmentation, while using CoMLP for both yields the best overall performance. Interestingly, the two settings exhibit different interaction preferences: inter-image fusion is more sensitive to removing the regional branch, suggesting the importance of local correspondence between spatially aligned anatomical and metabolic features, whereas vision-language fusion is more affected by removing the dilated branch, indicating a stronger reliance on global visual context for associating report-derived semantics with image regions. These observations suggest that a unified fusion operator need not impose identical interaction behavior across modalities; instead, complementary regional and dilated interactions allow the same cooperative gating principle to adapt to different characteristics of heterogeneous information.

Several limitations remain. Although the experiments cover five datasets, 2D/3D images, diverse imaging modalities, and multiple anatomical regions, simultaneous inter-image and vision-language fusion is mainly validated on PET/CT with clinical reports, and its effectiveness on other modality combinations such as CT/MR or multi-parametric MRI remains to be investigated. In addition, the interaction region size \(R\) is manually specified, although the sensitivity analysis shows relatively stable performance over a reasonable range. The current vision-language implementation also relies on provided textual descriptions; robustness to different reporting styles, incomplete descriptions, and noisier real-world reports warrants further study. Future work may explore adaptive interaction configuration, more parameter-efficient implementations, and hybrid MLP-attention designs for broader heterogeneous medical information fusion.

\section{Conclusion}
\label{sec:conclusion}
In this study, we have investigated MLPs as dense cross-modal interaction operators for fine-grained information fusion in medical image segmentation. We proposed CoMLP, which combines cooperative cross-gating with regional and dilated MLP interactions to capture complementary local and global dependencies. The same interaction primitive is used for both inter-image and vision-language fusion within a unified multi-source architecture. Experiments on five medical image segmentation benchmarks demonstrate consistent improvements over existing multi-modal and language-guided medical image segmentation methods, while further analyses show that CoMLP enables effective high-resolution cross-modal interaction with favorable memory scalability and provides complementary benefits across heterogeneous visual and textual information sources. These results suggest that MLP-based dense interaction offers a promising alternative for fine-grained cross-modal information fusion in medical vision.

\section*{REFERENCES}
\vspace{-1.5em}
\bibliographystyle{IEEEtran}
\bibliography{reference}

\end{document}